\pdfoutput=1

\documentclass[11pt]{article}

\usepackage[final]{ACL2023}
\usepackage{graphicx}

\usepackage{times}
\usepackage{latexsym}

\usepackage[T1]{fontenc}

\usepackage[utf8]{inputenc}

\usepackage{microtype}

\usepackage{inconsolata}

\title{LLM-RM at SemEval-2023 Task 2: Multilingual Complex NER using XLM-RoBERTa}

\author{Rahul Mehta \\
  IIIT Hyderabad,India \\
  \texttt{rahul.mehta@research.iiit.ac.in} \\\And
  Vasudeva Varma \\
 IIIT Hyderabad,India \\
  \texttt{vv@iiit.ac.in} \\}

\begin{document}
\maketitle
\begin{abstract}
Named Entity Recognition(NER) is a task of recognizing entities at a token level in a sentence. This paper focuses on solving NER tasks in a multilingual setting for complex named entities.Our team, LLM-RM participated in the recently organized SemEval 2023 task, Task 2: MultiCoNER II,Multilingual Complex Named Entity Recognition. We approach the problem by leveraging cross-lingual representation provided by fine-tuning XLM-Roberta base model on datasets of all of the 12 languages provided - Bangla, Chinese, English, Farsi, French, German, Hindi, Italian, Portuguese, Spanish, Swedish and Ukrainian. 

%
%
\end{abstract}

\section{Introduction}
\label{intro}

Named Entity Recognition(NER) is the task of recognizing entities (e.g., person, location, organization) in a piece of text.

Most of the NER datasets like CoNLL2003 \cite{conll2003}  are focussed on high-resource languages like English and are specific to a given domain like news.
The SemEval-2023 Task 2, Multilingual Complex Named Entity Recognition (MultiCoNER II) \cite{multiconer2-report} contains NER datasets curated for 12 languages which are low resource and from other domains as well. The task also provided a large number of 30 classes and simulated errors being added to the test set to make it more challenging.
It contains the following  languages: English, Spanish, Dutch, Russian, Turkish, Portugues, Korean, Farsi, German, Chinese, Hindi, and Bangla.

With the advent of Deep Learning based models, Transformer models like BERT, and bidirectional LSTM based ELMo became state of the art.
BERT's multilingual counterpart, mBERT became state of the art in multilingual NER tasks.
Another model XLM-RoBERTa(XLM-R) \cite{xlm-r-model} has shown to outperform mBERT on low resource languages on the NER tasks.
XLM-R is suitable for our task as it is pre-trained for more than 100 languages including the 4 languages which we worked upon. Therefore, we have used XLM-RoBERTa-base transformer model and fine-tuned it on each of the given 12 languages.

\subsection{The Challenge}
We participated in the shared task named  MultiCoNER II, which is part of Semeval 2023.
The task consists of building a named entity recognition system in 12 languages. 
There are 2 tracks to the tasks, one is the monolingual track for each language and the other is a multilingual track consisting of all languages. The condition of the competition was that one can only use the prediction of the monolingual model in the monolingual track and cannot use it for the multilingual track

The task is a successor to the  2022 challenge of MultiCoNER shared task, \cite{multiconer-report} where the key challenges were the following 1) syntactically complex entities like movies titles e.g Shawshank Redemption 2) these complex entities having low context and 3) long-tail entity distributions.

These challenges of NER for recognizing complex entities and in low-context situations were mentioned by \citet{meng2021gemnet}. The authors mention that for complex entities, some particular types (e.g creative works) can be linguistically complex. They can be complex noun phrases (Eternal Sunshine of the Spotless Mind), gerunds (Saving Private Ryan), infinitives (To Kill a Mockingbird), or full clauses (Mr Smith Goes to Washington). For long-tail entities, they mention that many domain entities have large long-tail distribution, with potentially millions of values (e.g., location names). It's very hard to build representative training data, as it can only cover a portion of the potentially infinite possible values.

The current MultiCoNER II challenge, expands the previous challenge of 2022 with new tasks and it emphasize the shortcomings of the current top models that are transformers based and depends on knowledge bases. It focusses on challenges like out of knowledge-base entities and noisy scenarios like the presence of spelling mistakes and typos.

\section{Related Work}
The earliest use of the neural networks to solve NER was by \cite{lstm-ner} who attempted to use an LSTM network to solve the problem. Recent transformer networks like BERT \cite{bert-ner} and ELMo \cite{elmo-ner} have further led to the state of the art results in the NER task for English language.

CRF layer \cite{crf} was also proposed to be used as a final layer for token classification. 

To expand NER in a multilingual setting, \cite{fetahu2021gazetteer} introduced a multilingual and code-mixed dataset concerning the search domain. They also used an XLM-RoBERTa model combined with gazetteers to build their multilingual NER classification system.

In a paper for the first edition of this task in 2022, XLM-RoBERTa has been applied to Hindi and English dataset and have shown to perform better than mBERT with a similar set of languages as in this current task \cite{silpa-nlp}.

\section{Data}
The dataset was first released in the Semeval task in 2022, called MultiCoNER dataset \cite{multiconer-data}.
In the 2nd edition of the task, The organisers provided a new dataset called MultiCoNER v2 \cite{multiconer2-data} comprising of individual language datasets in 12 languages.

Table \ref{tab:table1} contains the number of sentences in the training, development and test datasets per language in MulticoNER v2 dataset.

The test dataset is used for the final evaluation of the leaderboard and is further split into corrupted and non-corrupted sets. It is to be noted that the uncorrupted test set size is quite large compared to the training set for all the languages.

 
\begin{table}[h]
    \begin{tabular}{  |c | c | c | c | c | } 
      \hline
      Language & Train & Dev & Test-1 & Test-2 \\ 
      \hline
      BN & 9708 & 507 & 0 & 19,859 \\
      \hline
      ZH & 9,759 & 506 & 5696 & 14,569\\
      \hline 
      EN & 16,778 & 871 & 74,960 & 21,0267 \\
      \hline 
      DE & 9,785 & 512 & 5,880 & 16,334\\
      \hline 
      FA & 16,321 &	855 & 0 & 219,168\\ 
      \hline 
      FR & 16,548 & 857 & 74918 & 174,868 \\
      \hline
      HI & 9,632 & 514 & 0 &  18,406 \\
      \hline
      IT & 16,579	& 858 & 74,334 &  173,547\\
      \hline
      PT &  16,469 & 854 & 68,822 & 160,668 \\
      \hline 
      ES & 16,453 & 854	& 74,050 & 252,257 \\
      \hline
      SV & 16,363 & 856 & 69,342 & 161,848 \\
      \hline 
      UK & 16,429 & 851 & 0 & 238,296  \\
      \hline 
    \end{tabular}
    \caption{Sentences per split (Train,Test,Test-1 : Test-Corrupted, Test-2: Test-Uncorrupted) per language where BN is Bangla, ZH is Chinese, EN is English, FA is Farsi, FR is French, DE is German,HI is Hindi, IT is Italian, PT is Portuguese, ES is Spanish and UK is Ukranian language}
    \label{tab:table1}
\end{table}

Table \ref{tab:table2} contains the list of 30 entities present across all datasets and their grouping into the corresponding entity types.

\begin{table}[h]
    \begin{tabular}{ | c | c |  } 
      \hline
      Entity Name & Entity Type \\
      \hline
      Facility, OtherLOC,  & \\
      HumanSettlement, &  \\
      Station &  Location \\
      \hline
      VisualWork,  &\\
      MusicalWork, &\\
      WrittenWork, ArtWork, &\\
      Software & Creative Work \\
      \hline 
      MusicalGRP,& \\
      PublicCORP, & \\
      PrivateCORP, ORG, &\\
      AerospaceManufacturer,  &\\
      SportsGRP, &\\
      CarManufacturer   & Group\\
       \hline
     Scientist, Artist, & \\
     Athlete, OtherPER &\\
     Politician, Cleric, & \\
     SportsManager,  & Person\\ 
      \hline
      Clothing, Vehicle, &\\
      Food, Drink, OtherPROD  & Product\\
      \hline 
      Medication/Vaccine,  &\\
      MedicalProcedure, &\\
      AnatomicalStructure, &\\
      Symptom,& \\
      Disease       & Medical\\
      \hline 
     \end{tabular}
     \caption{List of Entities}
     \label{tab:table2}
\end{table}


\section{Methodology}

We utilised XLM-RoBERTa model for all 12 languages we participated in. 
XLM-RoBERTa is a massive Transformer trained on 100+ languages on 2TBs of CommonCrawl data.

\begin{figure}[]
\centerline{\includegraphics[scale=0.35]{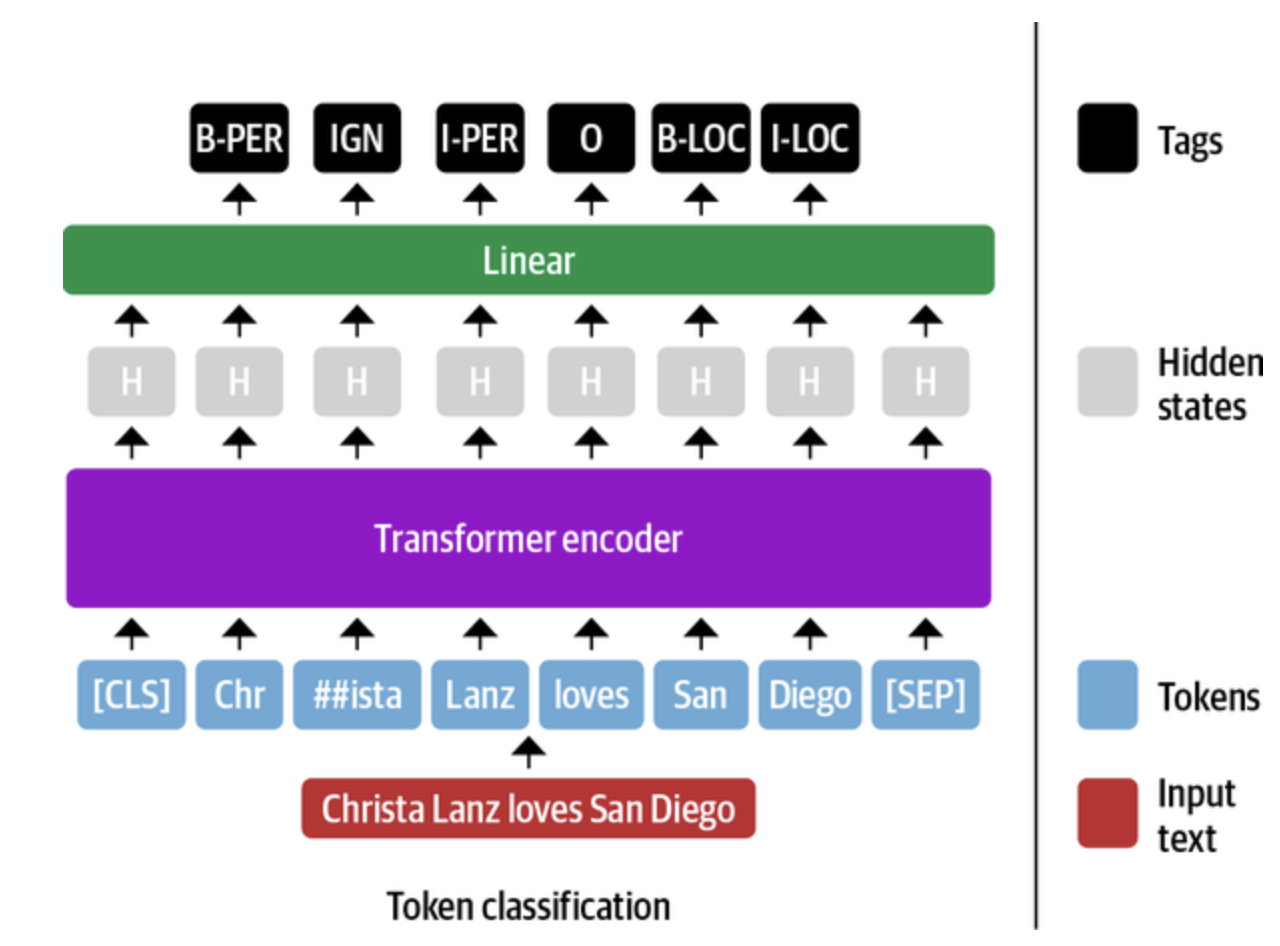}}
\caption{NER token classification using transformer encoder models like RoBERTa}
\label{fig:xlm}
\end{figure}

Figure \ref{fig:xlm} describes the architecture of transformer encoder models like RoBERTa.
It is trained with a multilingual MLM objective using only monolingual data.
XLM-RoBERTa has already been shown to outperform mBERT on cross-lingual classification by upto  23 percent accuracy on low-resource languages. It also gave competitive results with respect to the state of the art monolingual models.

\subsection{Experimental Set up}
We have used the PyTorch framework and HuggingFace's Tranformers library for our system.The training is done on a single GPU Nvidia Titan 2080 machine.

We used the XLM-RoBERTa-base model and fine-tuned it on each of the language datasets. We train the model in a batch size of 4 on the training dataset.
A dropout of 0.2 is applied between the last hidden layer and the output layer to prevent the model from overfitting. We used a learning rate of 2e-5.  For the optimizer, we used AdamW \cite{adamw} optimizer with an epsilon of 1e-08. AdamW is a stochastic optimization method and it modifies the implementation of weight decay in Adam, by separating weight decay from the gradient update. The max length of the sequence used is 16. We trained the model on 15 epochs and select the best one based on f1-scores on the test set.

For token classification at the end, we used a linear layer as the last layer to classify into the given 30 named entities.

The final model is based on the checkpoints of the model which are  selected based on best F1-scores on the development set.

\section{Results}
\label{sec:results}

Table \ref{tab:table3} shows the macro average scores of precision, recall and F-1.
\begin{table}[h]
    \begin{tabular}{ |c | c | c | c |} 
      \hline
      Language & Precision & Recall  & F1-Score\\ 
      \hline 
      BN & 62.18 & 61.60 & 60.46 \\
      \hline
      ZH & 26.51 & 27.34 & 25.50 \\ 
      \hline
      EN & 55.97 & 52.85  & 53.11\\
      \hline 
      FA & 53.41 & 54.87 & 53.13 \\ 
      \hline 
      FR & 53.17 & 54.42 & 53.33 \\
      \hline 
      DE & 66.56 & 59.92 & 61.62\\
      \hline
      HI & 70.64 & 69.17 & 69.04 \\
      \hline 
      IT & 60.48 & 62.48 & 60.97 \\ 
      \hline 
      PT & 64.08 & 63.59 & 63.41 \\ 
      \hline 
      ES & 61.02 & 58.19 & 58.32 \\
      \hline 
      SV & 57.19 & 60.03 & 57.55 \\
      \hline 
      UK & 55.59 & 49.22 & 49.06 \\ 
      \hline
    \end{tabular}
    \caption{Scores of XLM-RoBERTa-base on 12 languages on development set where BN is Bangla, ZH is Chinese, EN is English, FA is Farsi, FR is French, DE is German,HI is Hindi, IT is Italian, PT is Portuguese, ES is Spanish and UK is Ukranian language}
    \label{tab:table3}
\end{table}

\begin{table}[h]
    \begin{tabular}{ |c | c | c | c | c|} 
      \hline
      Entity-Type & HI & EN  & DE & SP\\ 
      \hline
      Group & 73.11 & 55.54 & 65.35 & 61.82\\
       \hline
      Medical & 75.69 & 56.27  & 55.18 & 62.68\\
       \hline
      Person & 58.87 & 46.99 & 51.80 & 48.66\\
       \hline
      Creative Work & 74.37 & 59.80 & 66.57 & 64.12 \\
      \hline
      Product & 59.95 & 46.88 & 48.90 & 54.55 \\
      \hline 
      Location& 76.15 & 55.08 & 64.17 & 55.86 \\
      \hline
    \end{tabular}
    \caption{F1-Scores of XLM-RoBERTa-base on 4 languages by Entity types on the development set}
    \label{tab:table4}
\end{table}
From Table \ref{tab:table4}, we observe that the entities belonging to the Creative Work category consistently have the highest F1-score across all 4 languages. Also, it can be noted that the entities belonging to Person and Product group have the lowest F1-scores for the model. 

\begin{table}[h]
    \begin{tabular}{ |c | c | c | c |} 
      \hline
      Lang & Overall F1 & Corrupted F1  & Uncor. F1\\ 
      \hline
      HI & 63.29 &  0 & 63.29 \\
       \hline
      EN & 52.08 & 46.3  & 54.73\\
       \hline
      DE & 55.54 & 54.1 & 54.73\\
       \hline
      ES & 54.81 & 49.32 & 57.42 \\
      \hline
    \end{tabular}
    \caption{Scores of XLM-RoBERTa-base on 4 languages on the final test set where HI is Hindi, EN is English,DE is German and ES is Spanish language}
    \label{tab:table5}
\end{table}
Table \ref{tab:table5} shows the macro average F1-scores of the final test set used for leaderboard 

From table \ref{tab:table3}  and table \ref{tab:table5}, we observe that the performance for Hindi drops most from development to test set, while for English and Spanish, it just drops slightly.

Also, we couldn't submit results for languages other than Hindi,English,German and Spanish by the time the competition ended, therefore we only have F1,corrupted f1 and uncorrupted F1 scores for these languages for the test set.


\section{Conclusion}
In this paper, we presented using XLM-RoBERTa-base to solve the shared task of MultiCoNER. 

Future work can include exploring more recent transformer-based models like XLM-V with very large vocabularies. Also, data augmentation techniques like entity replacement can be tried.


\bibliography{anthology,custom}
\bibliographystyle{acl_natbib}



\end{document}